\documentclass[11pt,a4paper]{article}
\usepackage[hyperref]{emnlp2018}
\usepackage{times}
\usepackage{latexsym}

\usepackage{url}

\aclfinalcopy 

\usepackage{graphicx}

\usepackage{array}
\usepackage{makecell}

\usepackage{multirow}
\usepackage{booktabs}

\usepackage{enumitem}
\usepackage{graphicx}

\usepackage{amsfonts}
\usepackage{amsmath}
\usepackage{amssymb}

\usepackage{blindtext}
\usepackage{scrextend}

\usepackage{threeparttable}

\usepackage{tabularx} 
\usepackage{float} 

\usepackage{mwe}

\usepackage{booktabs} 

\usepackage{diagbox} 

\usepackage{color, colortbl}  

\title{TVQA: Localized, Compositional Video Question Answering}

\author{
  Jie Lei $\;\;\;\;\;$ Licheng Yu $\;\;\;\;\;$ 
  Mohit Bansal $\;\;\;\;\;$ Tamara L. Berg \\
  Department of Computer Science \\ University of North Carolina at Chapel Hill \\
  {\tt \{jielei, licheng, mbansal, tlberg\}@cs.unc.edu} \\
  }
\date{}

\begin{document}
\maketitle
\begin{abstract}
Recent years have witnessed an increasing interest in image-based question-answering (QA) tasks.
However, due to data limitations, there has been much less work on video-based QA.
In this paper, we present TVQA, a large-scale video QA dataset based on 6 popular TV shows. TVQA consists of 152,545 QA pairs from 21,793 clips, spanning over 460 hours of video.
Questions are designed to be compositional in nature, requiring systems to jointly localize relevant moments within a clip, comprehend subtitle-based dialogue, and recognize relevant visual concepts.
We provide analyses of this new dataset as well as several baselines and a multi-stream end-to-end trainable neural network framework for the TVQA task. The dataset is publicly available at \url{http://tvqa.cs.unc.edu}.
\end{abstract}

\section{Introduction}

\begin{figure*}[ht!]
  \includegraphics[width=0.95\textwidth]{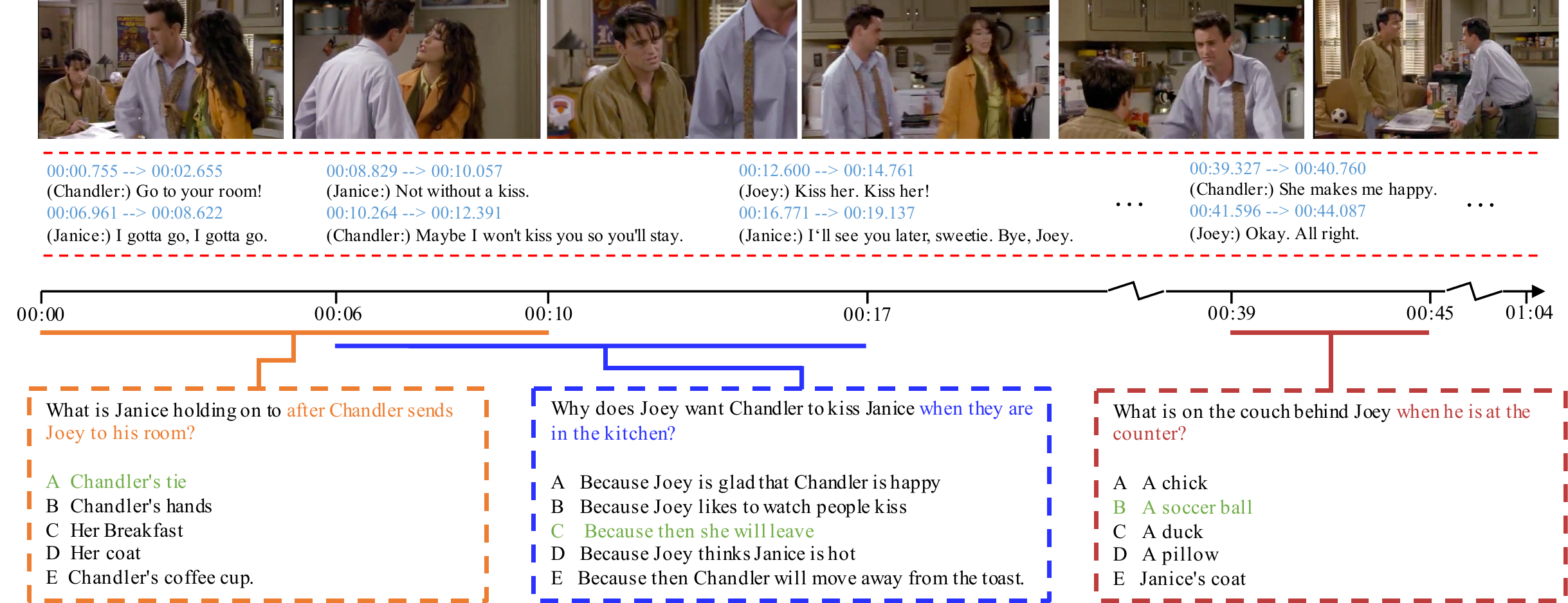}
  \vspace{-10pt}
  \caption{Examples from the TVQA dataset. All questions and answers are attached to 60-90 seconds long clips. For visualization purposes, we only show a few of the most relevant frames here. As illustrated above, some questions can be answered using subtitles or videos alone, while some require information from both modalities.}
  \label{fig:qa_examples_main}
  \vspace{-10pt}
\end{figure*}

Now that algorithms have started to produce relevant and realistic natural language that can describe images and videos, we would like to understand what these models truly comprehend. The Visual Question Answering (VQA) task provides a nice tool for fine-grained evaluation of such multimodal algorithms. VQA systems take as input an image (or video) along with relevant natural language questions, and produce answers to those questions. By asking algorithms to answer different types of questions, ranging from object identification, counting, or appearance, to more complex questions about interactions, social relationships, or inferences about why or how something is occurring, we can evaluate different aspects of a model's multimodal semantic understanding.

As a result, several popular image-based VQA datasets have been introduced, including DAQUAR~\cite{malinowski2014multi}, COCO-QA~\cite{ren2015exploring}, FM-IQA~\cite{gao2015you}, Visual Madlibs~
\cite{Yu2015VisualMF}, VQA~\cite{Antol2015VQAVQ}, Visual7W~\cite{zhu2016visual7w}, etc.
In addition, multiple video-based QA datasets have also been collected recently, e.g., MovieQA~\cite{Tapaswi2016MovieQAUS}, MovieFIB~\cite{maharaj2017dataset}, PororoQA~\cite{Kim2017DeepStoryVS}, TGIF-QA~\cite{Jang2017TGIFQATS}, etc.
However, there exist various shortcomings for each such video QA dataset.
For example, MovieFIB's video clips are typically short ($\sim$4 secs), and focused on purely visual concepts (since they were collected from audio descriptions for the visually impaired); MovieQA collected QAs based on text summaries only, making them very plot-focused and less relevant for visual information; PororoQA's video domain is cartoon-based; and TGIF-QA used pre-defined templates for generation on short GIFs.

With video-QA in particular, as opposed to image-QA, the video itself often comes with associated natural language in the form of (subtitle) dialogue. We argue that this is an important area to study because it reflects the real world, where people interact through language, and where many computational systems like robots or other intelligent agents will ultimately have to operate. As such, systems will need to combine information from what they see with what they hear, to pose and answer questions about what is happening.

We aim to provide a dataset that merges the best qualities from all of the previous datasets as well as focus on multimodal compositionality. In particular, we collect a new large-scale dataset that is built on natural video content with rich dynamics and realistic social interactions, where question-answer pairs are written by people observing both videos and their accompanying dialogues, encouraging the questions to require both vision and language understanding to answer. To further encourage this multimodal-QA quality, we ask people to write compositional questions consisting of two parts, a main question part, e.g. ``What are Leonard and Sheldon arguing about'' and a grounding part, e.g. ``when they are sitting on the couch''. This also leads to an interesting secondary task of QA temporal localization.

Our contribution is the TVQA dataset, built on 6 popular TV shows spanning 3 genres: medical dramas, sitcoms, and crime shows. On this data, we collected 152.5K human-written QA pairs (examples shown in Fig.\ref{fig:qa_examples_main}).
There are 4 salient advantages of our dataset.
First, it is large-scale and natural, containing 21,793 video clips from 925 episodes. On average, each show has 7.3 seasons, providing long range character interactions and evolving relationships. 
Each video clip is associated with 7 questions, with 5 answers (1 correct) for each question.
Second, our video clips are relatively long (60-90 seconds), thereby containing more social interactions and activities, making video understanding more challenging.
Third, we provide the dialogue (character name + subtitle) for each QA video clip.
Understanding the relationship between the provided dialogue and the question-answer pairs is crucial for correctly answering many of the collected questions.
Fourth, our questions are compositional, requiring algorithms to localize relevant moments (START and END points are provided for each question).

With the above rich annotation, our dataset supports three tasks: QA on the grounded clip, question-driven moment localization, and QA on the full video clip.
We provide baseline experiments on both QA tasks and introduce a state-of-the-art language and vision-based model (leaving moment localization for future work).

\section{Related Work}

\noindent\textbf{Visual Question Answering:}
Several image-based VQA datasets have recently been constructed, e.g., DAQUAR~\cite{malinowski2014multi}, VQA~\cite{Antol2015VQAVQ}, COCO-Q~\cite{ren2015exploring}, FM-IQA~\cite{gao2015you}, Visual Madlibs~\cite{Yu2015VisualMF}, Visual7W~\cite{zhu2016visual7w}, CLEVR~\cite{johnson2017clevr}, etc.
Additionally, several video-based QA datasets have also been proposed, e.g.
TGIF-QA~\cite{Jang2017TGIFQATS}, MovieFIB~\cite{maharaj2016dataset}, VideoQA~\cite{zhu2017uncovering}, LSMDC~\cite{rohrbach2015dataset}, TRECVID~\cite{over2014trecvid}, MovieQA~\cite{Tapaswi2016MovieQAUS}, PororoQA~\cite{Kim2017DeepStoryVS} and MarioQA~\cite{mun2017marioQA}. However, none of these datasets provides a truly realistic, multimodal QA scenario where both visual and language understanding are required to answer a large portion of questions, either due to unrealistic video sources (PororoQA, MarioQA) or data collection strategy being more focused on either visual (MovieFIB, VideoQA, TGIF-QA) or language (MovieQA) sources.
In comparison, our TVQA collection strategy takes a directly multimodal approach to construct a large-scale, real-video dataset by letting humans ask and answer questions while watching TV-show videos with associated dialogues.

\noindent\textbf{Text Question Answering:}
The related task of text-based question answering has been extensively explored~\citep{Richardson2013MCTestAC, Weston2015TowardsAQ, Rajpurkar2016SQuAD10, Hermann2015TeachingMT, Hill2015TheGP}. \citet{Richardson2013MCTestAC} collected MCTest, a multiple choice QA dataset intended for open-domain reading comprehension. With the same goal in mind, ~\citet{, Rajpurkar2016SQuAD10} introduced the SQuAD dataset, but their answers are specific spans from long passages.  ~\citet{Weston2015TowardsAQ} designed a set of tasks with automatically generated QAs to evaluate the textual reasoning ability of artificial agents and \citet{Hermann2015TeachingMT, Hill2015TheGP} constructed the cloze dataset on top of an existing corpus. While questions in these text QA datasets are specifically designed for language understanding, TVQA questions require both vision understanding and language understanding. 
Although methods developed for text QA are not directly applicable to TVQA tasks, they can provide inspiration for designing suitable models.

\noindent\textbf{Natural Language Object Retrieval:}
Language grounding addresses the task of object or moment localization in an image or video from a natural language description.
For image-based object grounding, there has been much work on phrase grounding~\cite{plummer2015flickr30k, wang2016learning, rohrbach2016grounding} and referring expression comprehension~\cite{Hu2016NaturalLO, yu2016modeling, nagaraja2016modeling, yu2017joint, Yu2018MAttNetMA}.
Recent work~\citep{Vasudevan2018ObjectRI} extends the grounding task to the video domain. 
Most recently, moment localization was proposed in~\cite{Hendricks2017LocalizingMI, Gao2017TALLTA}, where the goal is to localize a short moment from a long video sequence given a query description. 
Accurate temporal grounding is a necessary step to answering our compositional questions.

\section{TVQA Dataset}

\subsection{Dataset Collection}

We collected our dataset on 6 long-running TV shows from 3 genres: 1) sitcoms: \textit{The Big Bang Theory, How I Met Your Mother, Friends}, 2) medical dramas: \textit{Grey's Anatomy, House}, 3) crime drama: \textit{Castle}.
There are in total 925 episodes spanning 461 hours.
Each episode was then segmented into short clips. We first created clips every 60/90 seconds, then shifted temporal boudaries to avoid splitting subtitle sentences between clips.
Shows that are mainly conversational based, e.g., \textit{The Big Bang Theory}, were segmented into ~60 seconds clips, while shows that are less cerebral, e.g. \textit{Castle}, were segmented into ~90 seconds clips.
In the end, 21,793 clips were prepared for QA collection, accompanied with subtitles and aligned with transcripts to add character names.
A sample clip is shown in Fig.~\ref{fig:qa_examples_main}.

Amazon Mechanical Turk was used for VQA collection on video clips, where workers were presented with both videos and aligned named subtitles, to encourage multimodal questions requiring both vision and language understanding to answer.
Workers were asked to create questions using a compositional-question format: [What/How/Where/Why/...] $\rule{1cm}{0.15mm}$ [when/before/after] $\rule{1cm}{0.15mm}$.
The second part of each question serves to localize the relevant video moment within a clip, while the first part poses a question about that moment. This compositional format also serves to encourage questions that require both visual and language understanding to answer, since people often naturally use visual signals to ground questions in time, e.g. \textit{What was House saying before he leaned over the bed?} During data collection, we only used prompt words (when/before/after) to encourage workers to propose the desired, complex compositional questions. There were no additional template constraints. Therefore, most of the language in the questions is relatively free-form and complex.

Ultimately, workers pose 7 different questions for each video clip.
For each question, we asked workers to annotate the exact video portion required to answer the question by marking the START and END timestamps as in~\citet{Krishna2017DenseCaptioningEI}. In addition, they provide 1 correct and 4 wrong answers for each question.
Workers get paid \$1.3 for a single video clip annotation. The whole collection process took around 3 months.

To ensure the quality of the questions and answers, we set up an online checker in our collection interface to verify the question format, allowing only questions that reflect our two-step format to be submitted. The collection was done in batches of ~500 videos. For each harvested batch, we sampled 3 pairs of submitted QAs from each worker and checked the semantic correctness of the questions, answers, and timestamps. 

\begin{table}[t]
\centering
\scalebox{0.85}{
\begin{tabular}{ccccc}
\hline
QType & \#QA & Q. Len. & CA. Len. & WA. Len. \\ \hline
what & 84768 & 13.3 & 4.9 & 4.3 \\
who & 17654 & 13.4 & 3.1 & 3.0 \\
where & 17777 & 12.5 & 5.2 & 4.8 \\
why & 15798 & 14.5 & 9.0 & 7.7 \\
how & 13644 & 14.4 & 5.7 & 5.1 \\
others & 2904 & 15.2 & 4.9 & 4.7 \\ \hline
total & 152545 & 13.5 & 5.2 & 4.6 \\ \hline
\end{tabular}
}
\caption{Statistics for different question types based on first question word. Q = question, CA = correct answer, WA = wrong answer. Length is defined as the number of words in the sentence.}
\label{tab:qtype_stat}
\end{table}
 
\begin{figure}[t]
  \includegraphics[width=\linewidth]{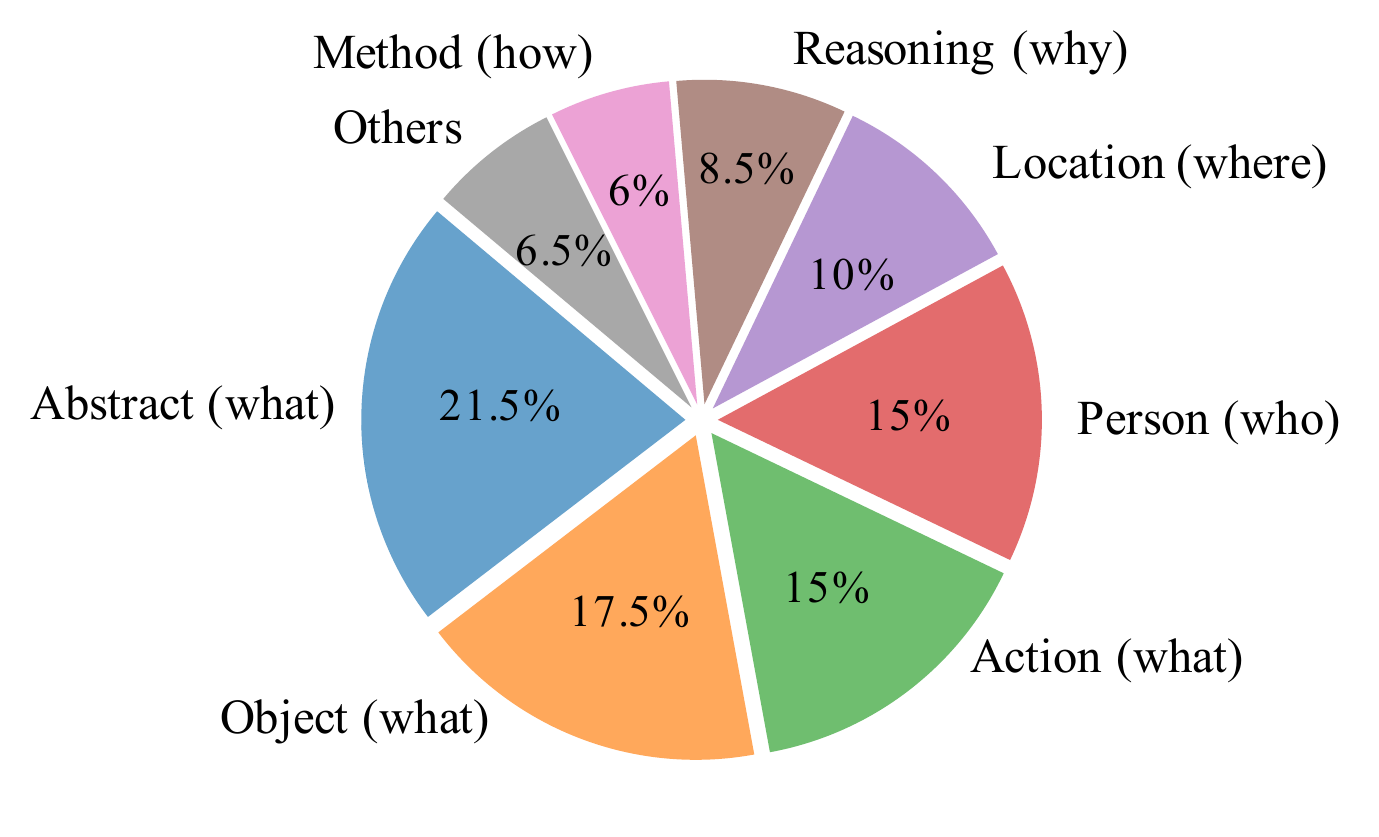}
  \vspace{-20pt}
  \caption{Distribution of question types based on answer types.}
  \label{fig:qtype_pie}
  \vspace{-4pt}
\end{figure}

\subsection{Dataset Analysis}

\textbf{Multiple Choice QAs:} Our QAs are multiple choice questions with 5 candidate answers for each question, for which only one is correct. 
Table~\ref{tab:qtype_stat} provides statistics of the QAs based on the first question word. On average, our questions contain 13.5 words, which is fairly long compared to other datasets. In general, correct answers tend to be slightly longer than wrong answers. 
Fig.~\ref{fig:qtype_pie} shows the distribution of different questions types.
Note ``what'' (Abstract, Object, Action), ``who'' (Person), ``why'' (Reasoning) and ``where'' (Location) questions form a large part of our data.

The negative answers in TVQA are written by human annotators. They are instructed to write false but relevant answers to make the negatives challenging. 
Alternative methods include sampling negative answers from other questions' correct answers, either based on semantic similarity ~\citep{Das2017VisualD, Jang2017TGIFQATS} or randomly~\citep{Antol2015VQAVQ, Das2017VisualD}. 
The former is prone to introducing paraphrases of the ground-truth answer~\citep{zhu2016visual7w}.  The latter avoids the problem of paraphrasing, but generally produces irrelevant negative choices. We show in Table~\ref{tab:rand_negative} that our human written negatives are more challenging than randomly sampled negatives.

\begin{figure}[ht]
\centering
  \includegraphics[width=0.85\linewidth]{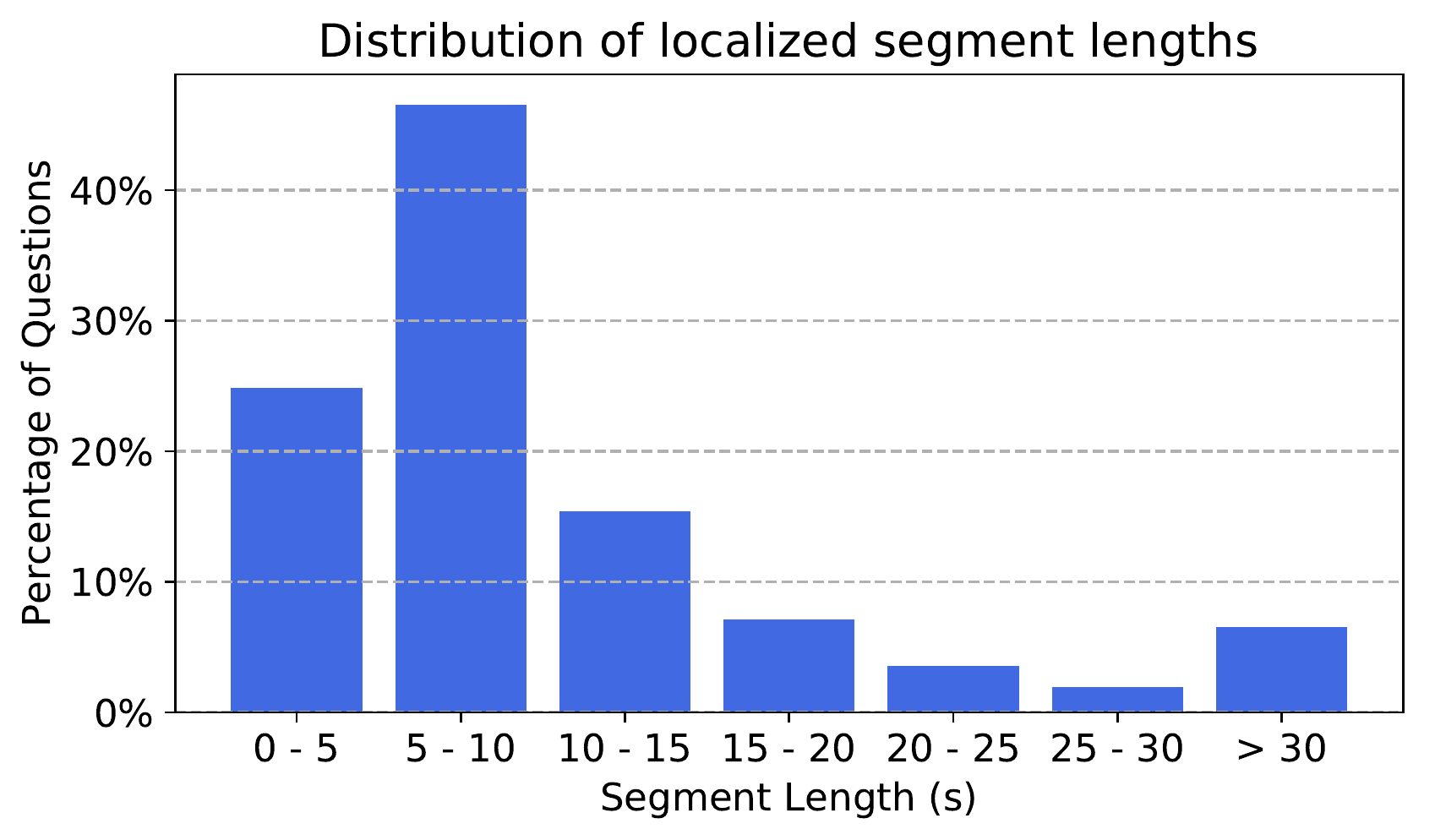}
  \vspace{-10pt}
  \caption{Distribution of localized segment lengths. The majority of our questions have timestamp localized segment with length less than 15 seconds.}
  \label{fig:ts_len_dist}
\end{figure}

\begin{table}[ht]
\small
\centering
\scalebox{0.96}{
\begin{tabular}{llrrrr}
\hline
Show    & Genre     & \#Sea. & \#Epi. & \#Clip & \#QA  \\ \hline
BBT     & sitcom    & 10        & 220        & 4,198   & 29,384 \\
Friends & sitcom    & 10        & 226        & 5,337   & 37.357 \\
HIMYM   & sitcom    & 5         & 72         & 1,512   & 10,584  \\
Grey    & medical   & 3         & 58         & 1,427   & 9,989  \\
House   & medical   & 8         & 176        & 4,621   & 32,345 \\
Castle  & crime & 8         & 173        & 4,698   & 32,886 \\ \hline
Total   & \quad--- & 44 & 925  &21,793 & 152,545 \\\hline
\end{tabular}
}
\vspace{-10pt}
\caption{Data Statistics for each TV show. BBT = \textit{The Big Bang Theory}, HIMYM = \textit{How I Met You Mother}, Grey = \textit{Grey's Anatomy}, House = \textit{House M.D.}, Epi = Episode, Sea. = Season}
\label{tab:dset_stat_by_genre}
\end{table}

\begin{table}[!ht]
\centering
\scalebox{0.75}{
\begin{tabular}{l|l}
\hline 
Show                & \multicolumn{1}{c}{Top unique nouns}                                                         \\ \hline
\multirow{2}{*}{BBT}     & game, mom, laptop, water, store, dinner, book,       \\
                         & stair, computer, food, wine, glass, couch, date         \\ \cline{2-2}
\multirow{2}{*}{Friends} & shop, kiss, hair, sofa, jacket, counter, coffee,             \\
                         & everyone, coat, chair, kitchen, baby, apartment               \\ \cline{2-2}
\multirow{2}{*}{HIMYM}   & bar, beer, drink, job, dad, sex, restaurant, wedding,            \\
                         & party, booth, dog, story,  bottle, club, painting        \\ \hline
\multirow{2}{*}{Grey}    & nurse, side, father, hallway, scrub, chart, wife,        \\
                         & window, life, family, chief, locker, head, surgery      \\ \cline{2-2}
\multirow{2}{*}{House}   & cane, team, blood, test, brain, pill, office,  pain,    \\
                         & symptom, diagnosis, hospital, coffee, cancer, drug          \\ \hline
\multirow{2}{*}{Castle}  & gun, victim, picture, case, photo, body, murder,       \\ 
                         & suspect, scene, crime, money, interrogation  \\ \hline
\end{tabular}
}
\caption{Top unique nouns in questions and correct answers.}
\label{tab:freq_words_by_genre}
\vspace{-10pt}
\end{table}

\begin{table*}[ht!]
\setcounter{table}{4}
\small
\centering
\scalebox{0.95}{
\begin{tabular}{llllllccc}
\hline
\multirow{2}{*}{Dataset}   & \multirow{2}{*}{V. Src.} & \multirow{2}{*}{QType}    & \multirow{2}{*}{\#Clips / \#QAs}   & Avg.& Total & \multicolumn{2}{c}{Q. Src.} & Timestamp \\
          &         &          &                   &  Len.(s)      & Len.(h)      & text         & video        &   annotation      
  \\ \hline
MovieFIB~\citep{maharaj2017dataset}  & Movie   & OE       & 118.5k / 349k & 4.1           & 135            &  \checkmark  &    -      &    -               \\
Movie-QA~\citep{Tapaswi2016MovieQAUS}  & Movie   & MC       & 6.8k / 6.5k     & 202.7         & 381            &  \checkmark  &    -         &  \checkmark         \\
TGIF-QA~\citep{Jang2017TGIFQATS}   & Tumblr  & OE\&MC & 71.7k / 165.2k  & 3.1           & 61.8           &   \checkmark     &  \checkmark     &    -                   \\
Pororo-QA~\citep{Kim2017DeepStoryVS} & Cartoon & MC       & 16.1k / 8.9k     & 1.4          & 6.3           &  \checkmark  &  \checkmark &     -                  \\ \hline
TVQA (our)      & TV show & MC       & 21.8k / 152.5k   & 76.2          & 461.2            &  \checkmark  &  \checkmark  &  \checkmark         \\ \hline
\end{tabular}
}
\vspace{-5pt}
\caption{Comparison of TVQA to various existing video QA datasets. OE = open-ended, MC = multiple-choices. Q. Src. = Question Sources, it indicates where the questions are raised from. TVQA dataset is unique since its questions are based on both text and video, with additional timestamp annotation for each of them. It is also significantly larger than previous  datasets in terms of total length of videos.}
\label{tab:dset_comparison}
\end{table*}

\begin{table}[t]
\setcounter{table}{3}
\centering
\scalebox{0.75}{
\begin{tabular}{l|l}
\hline 
Character                & \multicolumn{1}{c}{Top unique nouns}                                                         \\ \hline
\multirow{2}{*}{Sheldon}     & Arthur, train, Kripke, flag, flash, Wil,     \\
                         & logo, Barry, superhero, Spock, trek, sword        \\ \hline
\multirow{2}{*}{Leonard} & Leslie, helium, robe, Dr, team, Kurt             \\
                         & university, key, chess, Stephen              \\ \hline
\multirow{2}{*}{Howard}   & NASA, trick, van, language, summer,             \\
                         & letter, Mike, station, peanut, Missy       \\  \hline
\multirow{2}{*}{Raj}    & Lucy, Claire, parent, music, nothing,        \\
                         & Isabella, bowl, sign, back, India, number     \\  \hline
\multirow{2}{*}{Penny}   & basket, order, mail, mouth, cheesecake, factory   \\
                         & shower, pizza, cream, Alicia, waitress, ice \\ \hline
\multirow{2}{*}{Amy}  & Dave, meemaw, tablet, birthday, monkey, coat,       \\ 
                         & brain, ticket, laboratory, theory, lip, candle \\ \hline
\multirow{2}{*}{Bernadette}  & song, sweater, wedding, child,  husband,     \\ 
                         & everyone, necklace, stripper, weekend, airport \\ \hline
\end{tabular}
}
\caption{Top unique nouns for characters in BBT.}
\label{tab:bbt_cooccur}
\vspace{-5pt}
\end{table}

\noindent\textbf{Moment Localization:} 
The second part of our question is used to localize the most relevant video portion to answer the question. 
The prompt of ``when'', ``after'', ``before'' account for 60.03\%, 30.19\% and 9.78\% respectively of our dataset.
TVQA provides the annotated START and END timestamps for each QA.
We show the annotated segment lengths in Fig.~\ref{fig:ts_len_dist}.
We found most of the questions rely on relatively short moments (less than 15 secs) within a longer clip (60-90 secs).

\noindent\textbf{Differences among our 6 TV Shows:} The videos used in our dataset are from 6 different TV shows. Table~\ref{tab:dset_stat_by_genre} provides statistics for each show. A good way to demonstrate the difference among questions from TV shows is to show their top unique nouns. 
In Table~\ref{tab:freq_words_by_genre}, we present such an analysis. The top unique nouns in sitcoms (\textit{BBT}, \textit{Friends}, \textit{HIMYM}) are mostly daily objects, scenes and actions, while medical dramas (\textit{Grey, House}) questions contain more medical terms, and crime shows (\textit{Castle}) feature detective terms. Although similar, there are also notable differences among shows in the same genre. For example, \textit{BBT} contains ``game'' and ``laptop''  while \textit{HIMYM} contains ``bar'' and ``beer'', indicating the different major activities and topics in each show.
Additionally, questions about different characters also mention different words, as shown in Table~\ref{tab:bbt_cooccur}.

\noindent\textbf{Comparison with Other Datasets:}
Table~\ref{tab:dset_comparison} presents a comparison of our dataset to some recently proposed video question answering datasets. In terms of total length of videos, TVQA is the largest, with a total of 461.2 hours of videos. 
MovieQA~\citep{Tapaswi2016MovieQAUS} is most similar to our dataset, with both multiple choice questions and timestamp annotation. 
However, their questions and answers are constructed by people posing questions from a provided plot summary, then later aligned to the video clips, which makes most of their questions text oriented.

\begin{table}[t]
\small
\centering
\scalebox{0.95}{
\begin{tabular}{lc} 
\hline
VQA source & Test-Public accuracy.  \\ \hline
Question & 32.61  \\
Video and Question & 61.96  \\
Subtitle and Question & 73.03  \\
Video, Subtitle, and Question & 89.41 \\\hline
\end{tabular}
}
\caption{Human accuracy on test-public set based on different sources. As expected, humans get the best performance when given both videos and subtitles.}
\label{tab:human_acc}
\vspace{-5pt}
\end{table}

\noindent\textbf{Human Evaluation on Usefulness of Video and Subtitle in Dataset:} To gain a better understanding of the roles of videos and subtitles in the our dataset, we perform a human study, asking different groups of workers to complete the QA task in settings while observing different sources (subsets) of information:
\begin{itemize}[itemsep=1mm, parsep=0pt]
    \item Question only. 
    \item Video and Question.
    \item Subtitle and Question.
    \item Video, Subtitle, and Question.
\end{itemize}
\noindent We made sure the workers that have written the questions did not participate in this study and that workers see only one of the above settings for answering each question. 
Human accuracy on our test-public set under these 4 settings are reported in Table~\ref{tab:human_acc}. As expected, compared to human accuracy based only on question-answer pairs (Q), adding videos (V+Q), or subtitles (S+Q) significantly improves human performance. Adding both videos and subtitles (V+S+Q) brings the accuracy to 89.41\%. This indicates that in order to answer the questions correctly, both visual and textual understanding are essential.
We also observe that workers obtain 32.61\% accuracy given question-answer pairs only, which is higher than random guessing (20\%).
We ascribe this to people's prior knowledge about the shows. Note, timestamp annotations are not provided in these experiments.

\section{Methods}
\begin{figure*}[h!]
  \includegraphics[width=\textwidth]{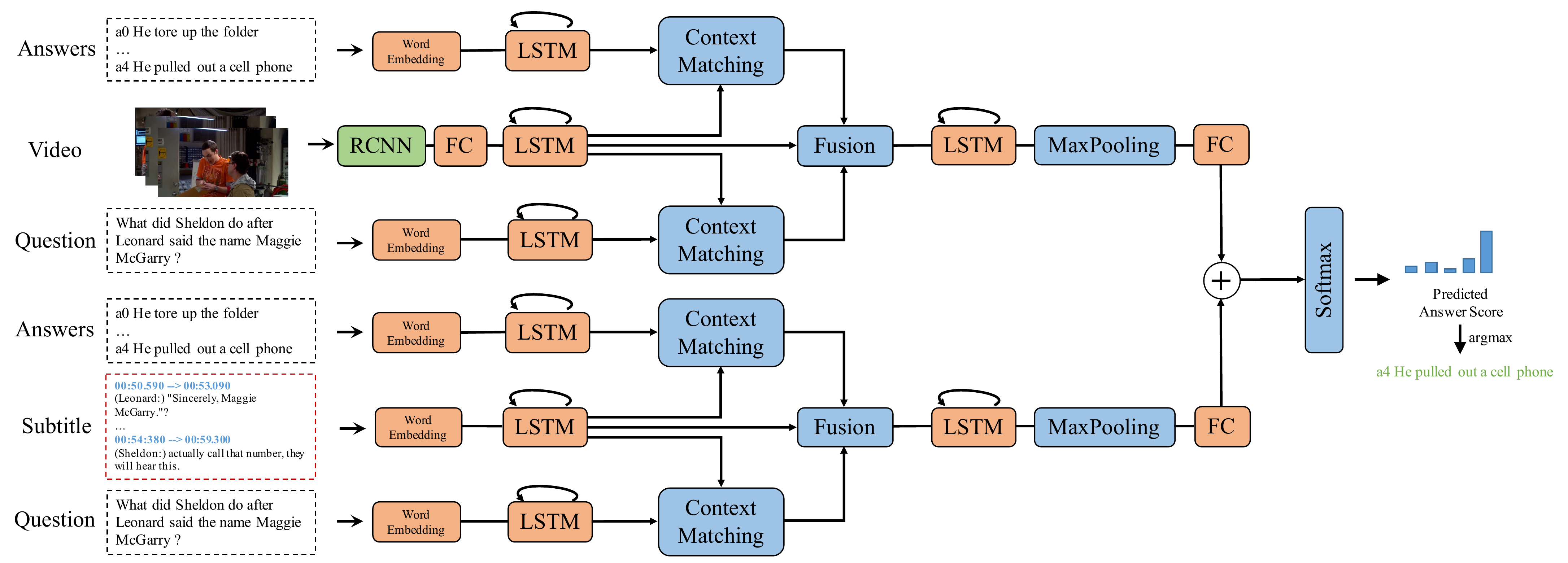}
  \vspace{-20pt}
  \caption{Illustration of our multi-stream model for Multi-Modal Video QA. Our full model takes different contextual sources (regional visual features, visual concept features, and subtitles) along with question-answer pair as inputs to each stream. For brevity, we only show regional visual features (upper) and subtitle (bottom) streams.}
  \label{fig:model_main}
  \vspace{-5pt}
\end{figure*}

We introduce a multi-stream end-to-end trainable neural network for Multi-Modal Video Question Answering. Fig.~\ref{fig:model_main} gives an overview of our model. Formally, we define the inputs to the model as: a 60-90 second video clip $V$,  a subtitle $S$, a question $q$, and five candidate answers $\{a_i\}_{i=0}^4$.

\subsection{Video Features}
Frames are extracted at 3 fps. We run Faster R-CNN~\cite{Ren2015FasterRT} trained on the Visual Genome~\cite{Krishna2017DenseCaptioningEI} to detect object and attribute regions in each frame.  Both regional features and predicted detection labels can be used as model inputs. We also use ResNet101~\citep{He2016DeepRL} trained on ImageNet~\citep{Deng2009ImageNetAL} to extract whole image features.

\noindent\textbf{Regional Visual Features:} On average, our videos contain ~229 frames, with 16 detections per frame. 
It is not trivial to model such long sequences.
For simplicity, we follow~\cite{Anderson2017BottomUpAT, karpathy2015deep} selecting the top-K regions\footnote{Based on cross-validation, we find K=6 to perform best.} from each detected label across all frames.
Their regional features are L2-normalized and stacked together to form our visual representation $V^{reg} \in \mathbb{R}^{n_{reg} \times 2048}$. 
Here $n_{reg}$ is the number of selected regions.

\begin{figure}[t]
  \includegraphics[width=\linewidth]{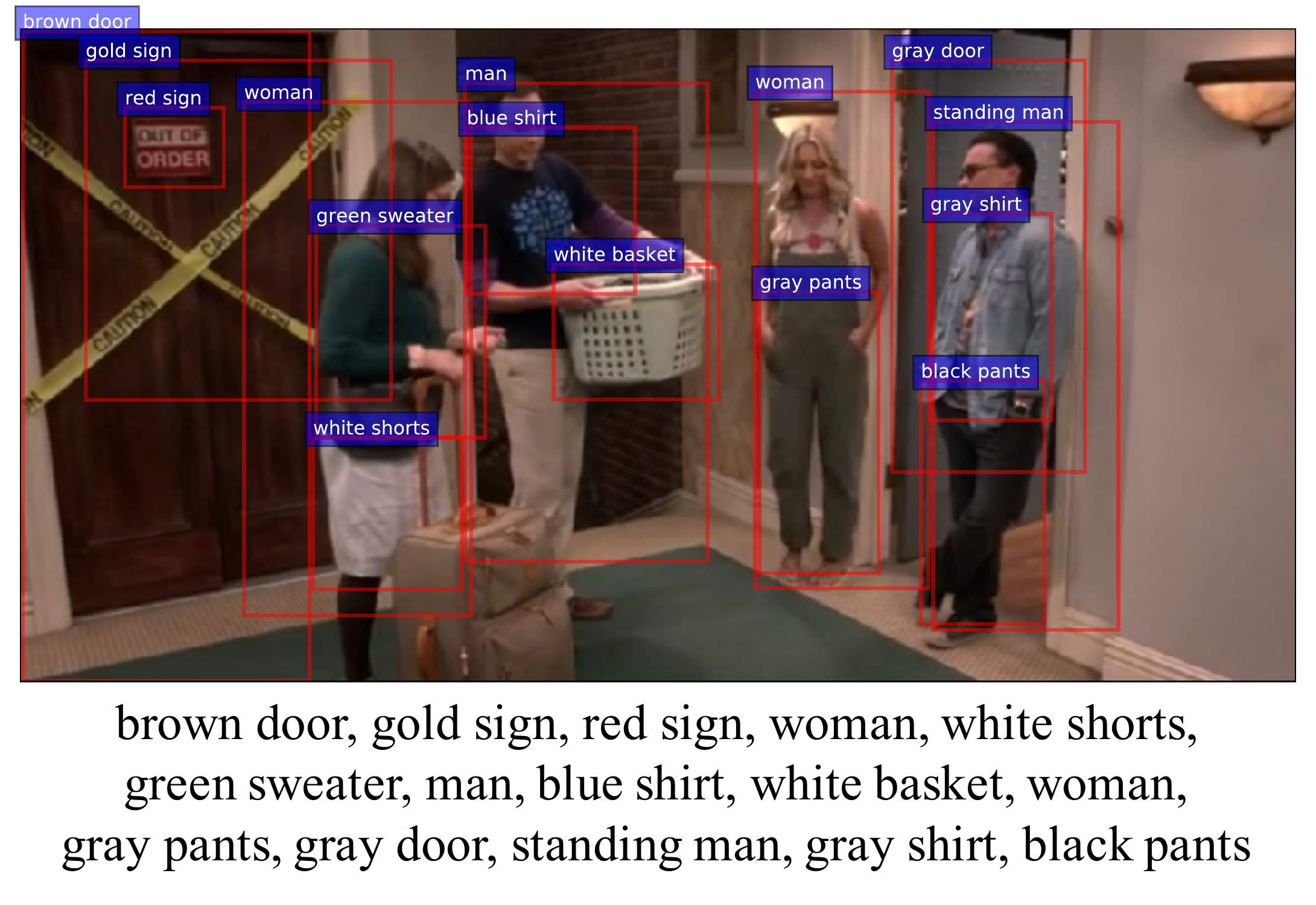}
  \vspace{-20pt}
  \caption{Faster R-CNN detection example. The detected object labels and attributes can be viewed as a description to the frame, which is potentially helpful to answer a visual question.}
  \label{fig:detection_example}
  \vspace{-5pt}
\end{figure}

\noindent\textbf{Visual Concept Features:} 
Recent work~\cite{yin2017obj2text} found that using detected object labels as input to an image captioning system gave comparable performance to using CNN features directly. Inspired by this work, we also experiment with using detected labels as visual inputs. As shown in Fig.~\ref{fig:detection_example}, we are able to detect rich visual concepts, including both objects and attributes, e.g. ''white basket'', which could be used to answer ``What is Sheldon holding in his hand when everyone is at the door''.
We first gather detected concepts over all the frames to represent concept presence. After removing duplicate concepts, we use GloVe~\citep{Pennington2014GloveGV} to embed the words. The resulting video representation is denoted as $V^{cpt} \in \mathbb{R}^{n_{cpt} \times 300}$, where $n_{cpt}$ is the number of unique concepts.

\noindent\textbf{ImageNet Features:} 
We extract the pooled 2048D feature of the last block of ResNet101. 
Features from the same video clip are L2 normalized and stacked, denoted as  $V^{img} \in \mathbb{R}^{n_{img} \times 2048}$, where $n_{img}$ is the number of frames extracted from the video clip.

\subsection{LSTM Encoders for Video and Text}
We use a bi-directional LSTM (BiLSTM) to encode both textual and visual sequences. 
A subtitle $S$, which contains a set of sentences, is flattened into a long sequence of words and GloVe~\citep{Pennington2014GloveGV} is used to embed the words. 
We stack the hidden states of the BiLSTM from both directions at each timestep to obtain the subtitle representation $H^{S} \in \mathbb{R}^{n_S \times 2d}$, where $n_S$ is the number of subtitle words, $d$ is the hidden size of the BiLSTM (set to 150 in our experiments). 
Similarly, we encode question $H^q \in \mathbb{R}^{n_q \times 2d}$, candidate answers $H^{a_i} \in \mathbb{R}^{n_{a_i} \times 2d}$, and visual concepts $H^{cpt} \in \mathbb{R}^{n_{cpt} \times 2d}$.  $n_q$ and $n_{a_i}$ are the number of words in question and answer $a_i$, respectively.
Regional features $V^{reg}$ and ImageNet features $V^{img}$ are first projected into word vector space using a non-linear layer with tanh activation, then encoded using the same BiLSTM to obtain the regional representations $H^{reg} \in \mathbb{R}^{n_{reg} \times 2d}$ and $H^{img} \in \mathbb{R}^{n_{img} \times 2d}$, respectively.

\subsection{Joint Modeling of Context and Query}

We use a context matching module and BiLSTM to jointly model the contextual inputs (subtitle, video) and query (question-answer pair).
The context matching module is adopted from the context-query attention layer from previous works~\citep{Seo2016BidirectionalAF, Yu2018QANetCL}.
It takes context vectors and query vectors as inputs and produces a set of context-aware query vectors based on the similarity between each context-query  pair. 

Taking the regional visual feature stream as an example (Fig.~\ref{fig:model_main} upper stream), where $H^{reg}$ is used as context input\footnote{For visual concept features and ImageNet features, we simply replace $H^{reg}$ with $H^{cpt}$ or $H^{img}$ as the context.}.
The question embedding, $H^{q}$, and answer embedding, $H^{a_i}$, are used as  queries. 
After feeding context-query pairs into the context matching module, we obtain a video-aware-question representation, $G^{reg,q} \in \mathbb{R}^{n_{reg} \times 2d}$, and video-aware-answer representation, $G^{reg,a_i} \in \mathbb{R}^{n_{reg} \times 2d}$, which are then fused with video context:
\begin{align*}
    M^{reg,a_i} = [&H^{reg}; G^{reg,q}; G^{reg,a_i}; \\
               &H^{reg} \odot G^{reg,q}; H^{reg} \odot G^{reg,a_i}],
\end{align*}
where $\odot$ is element-wise product.
The fused feature, $M^{reg,a_i} \in \mathbb{R}^{n_{reg} \times 10d}$, is fed into another BiLSTM.
Its hidden states, $U^{reg,a_i} \in \mathbb{R}^{n_{reg} \times 10d}$, are max-pooled temporally to get the final vector, $u^{reg,a_i} \in \mathbb{R}^{10d}$, for answer $a_i$.
We use a linear layer with softmax to convert $\{{u^{reg,a_i}}\}_{i=0}^4$ into answer probabilities.
Similarly, we can compute the answer probabilities given subtitle as context (Fig.~\ref{fig:model_main} bottom stream).
When multiple streams are used, we simply sum up the scores from each stream as the final score~\citep{Wang2016TemporalSN}.

\section{Experiments} \label{sec:experiments}

For evaluation, we introduce several baselines and compare them to our proposed model.  

In all experiments, setup is as follows.
We split the TVQA dataset into 80\% train, 10\% val, and 10\% test splits such that videos and their corresponding QA pairs appear in only one split. This results in 122,039 QA pairs for training, 15,253 QA pairs for validation, and 15,253 QA pairs for testing. We further split the test set into two subsets, test-public (7623 QA pairs) and test-reserved (7630 QA pairs), where test-public is used in our leaderboard\footnote{\url{http://tvqa.cs.unc.edu/leaderboard.html}}, test-reserved is reserved for future use. We evaluate each model using multiple-choice question answering accuracy.

\subsection{Baselines}

\noindent\textbf{Longest Answer:} Table~\ref{tab:qtype_stat} indicates that the average length of the correct answers is longer than the wrong ones; thus, our first baseline simply selects the longest answer for each question.

\noindent\textbf{Nearest Neighbor Search:} 
In this baseline, we use Nearest Neighbor Search (NNS) to compute the closest answer to our question or subtitle.
We embed sentences into vectors using TFIDF, SkipThought~\cite{Kiros2015SkipThoughtV}, or averaged GloVe~\citep{Pennington2014GloveGV} word vectors, then compute the cosine similarity for each question-answer pair or subtitle-answer pair.
For TFIDF, we use bag-of-words to represent the sentences, assigning a TFIDF value for each word.

\noindent\textbf{Retrieval:} 
Due to the size of TVQA, there may exist similar questions and answers in the dataset. Thus, we also implement a baseline two-step retrieval approach: 
given a question and a set of candidate answers, we first retrieve the most relevant question in the training set, then pick the candidate answer that is closest to the retrieved question's correct answer. 
Similar approaches have also been used in dialogue systems~\citep{Jafarpour2010FilterRA, Leuski2011NPCEditorCV}, picking the appropriate responses to an utterance from a predefined human conversational corpus. 
Similar to NNS, we use TFIDF, SkipThought, and GloVe vectors with cosine similarity.

\begin{table}[ht]
\centering
\scalebox{0.80}{
\begin{tabular}{l|l|c|cc}
\hline
   &                        & Video   & \multicolumn{2}{c}{\small{Test-Public Accuracy}}  \\ \cline{4-5}
   & Method                 & Feature & w/o ts & w/ ts \\ \hline \hline
0  & Random                 &  -     & 20.00                & 20.00             \\ \hline
1  & Longest Answer         &  -      & 30.22             & 30.22          \\ \hline
2  & Retrieval-Glove        &  -      & 22.77             & 22.77          \\
3  & Retrieval-SkipThought &  -      & 24.27             & 24.27          \\
4  & Retrieval-TFIDF       &  -      & 20.78             & 20.78          \\ \hline
5  & NNS-Glove Q         &  -      & 22.63             & 22.63          \\
 
6  & NNS-SkipThought Q         &  -      & 23.39             & 23.39          \\
7  & NNS-TFIDF Q                &  -      & 19.47             & 19.47          \\ \hline
8  & NNS-Glove S         &  -      & 23.74             & 29.95           \\
9  & NNS-SkipThought S         &  -      & 26.93             & 38.29           \\ 
10  & NNS-TFIDF S                &  -      & 49.59             & 50.79          \\ \hline \hline
11  & Our Q                  &  -      & 43.50             & 43.50          \\ \hline
12  & Our V+Q                & img     & 42.87             & 43.91          \\
13  & Our V+Q                & reg     & 43.01             & 45.23          \\
14 & Our V+Q                & cpt     & 43.84             & 45.44          \\ \hline
15 & Our S+Q                &  -      & 62.69             & 66.36          \\ \hline
16 & Our S+V+Q              & img  & 63.44           & 66.94          \\
17 & Our S+V+Q              & reg     & 63.06             & 68.19          \\
18 & Our S+V+Q              & cpt     & \textbf{66.46}    & \textbf{68.48} \\ \hline
\end{tabular}
}
\caption{
Accuracy for different methods on TVQA test-public set. Q = Question, S = Subtitle, V = Video, img = ImageNet features, reg = regional visual features, cpt = visual concept features, ts = timestamp annotation. Human performance without timestamp annotation is reported in Table~\ref{tab:human_acc}.
}
\label{tab:main_results}
\end{table}

\subsection{Results}

Table~\ref{tab:main_results} shows results from baseline methods and our proposed neural model. 
Our main results are obtained by using full-length video clips and subtitles, without using timestamps (\textit{w/o ts}). 
We also run the same experiments using the localized video and subtitle segment specified by the ground truth timestamps (\textit{w/ ts}). 
If not indicated explicitly, the numbers described below are from the experiments on full-length video clips and subtitles.

\noindent\textbf{Baseline Comparison:} Row 1 shows results of the longest answer baseline, achieving 30.22\% (compared to random chance at 20\%). As expected, the retrieval-based methods (row 2-4) and the answer-question similarity based methods (row 5-7) perform rather poorly, since no contexts (video or subtitle) are considered. 
When using subtitle-answer similarity to choose correct answers, Glove, SkipThought, and TFIDF based approaches (row 8-10) all achieve significant improvement over question-answer similarity. 
Notably, TFIDF (row 10) answers 49.59\% of the questions correctly. 
Since our questions are raised by people watching the videos, it is natural for them to ask questions about specific and unique objects/locations/etc., mentioned in the subtitle. Thus, it is not surprising that TFIDF based similarity between answer and subtitle performs so well.

\noindent\textbf{Variants of Our Model:} Rows 11-18 show results of our model with different contextual inputs and features. The model that only uses question-answer pairs (row 11) achieves 43.50\% accuracy. Compared to the subtitle model (row 15), adding video as additional sources (row 16-18) improves performance. 
Interestingly, adding video to the question only model (row 11) do not work as well (row 12-14).
Our hypothesis is that the video feature streams may be struggling to learn models for answering textual questions, which degrades their ability to answer visual questions. Overall, the best performance is achieved by using all the contextual sources, including subtitles and videos (using concept features, row 18).

\begin{table}[t]
\setlength\aboverulesep{0pt} \setlength\belowrulesep{0pt}
\centering
\scalebox{0.58}{
\begin{tabular}{c|cccccccc}
\hline
\multirow{2}{*}{} &\multirow{2}{*}{Q} & \multirow{2}{*}{S+Q} & \multicolumn{3}{c}{V+Q} & \multicolumn{3}{c}{S+V+Q} \\ \cmidrule(rl){4-6} \cmidrule(l){7-9}
 &  &  & img & reg & cpt & img & reg & cpt \\ \hline
what (55.62\%) & 44.11 & 62.29 & 44.96 & 45.93 & 47.44 & 63.88 & 65.28 & 66.05 \\
who (11.55\%) & 36.55 & 68.33 & 35.75 & 34.85 & 34.68 & 67.76 & 67.20 & 67.99 \\
where (11.67\%) & 42.58 & 56.97 & 47.13 & 48.43 & 48.20 & 61.97 & 63.71 & 61.46 \\
how (8.98\%) & 41.17 & 71.97 & 41.17 & 42.41 & 40.95 & 71.17 & 70.80 & 71.53 \\
why (10.38\%) & 45.23 & 78.65 & 46.05 & 45.36 & 45.48 & 78.33 & 77.13 & 78.77 \\
other (1.80\%) & 36.50 & 74.45 & 37.23 & 36.50 & 33.58 & 73.72 & 72.63 & 74.09 \\ \hline
all (100\%) & 42.77 & 65.15 & 43.78 & 44.40 & 45.03 & 66.44 & 67.17 & 67.70 \\ \hline
\end{tabular}
}
\caption{Accuracy of each question type using different models (w/ ts) on TVQA Validation set. Q = Question, S = Subtitle, V = Video, img = ImageNet features, reg = regional visual features, cpt = visual concept features. The percentage of each question type is shown in brackets.}
\label{tab:qtype_acc}
\end{table}

\begin{figure*}[t]
  \includegraphics[width=\textwidth]{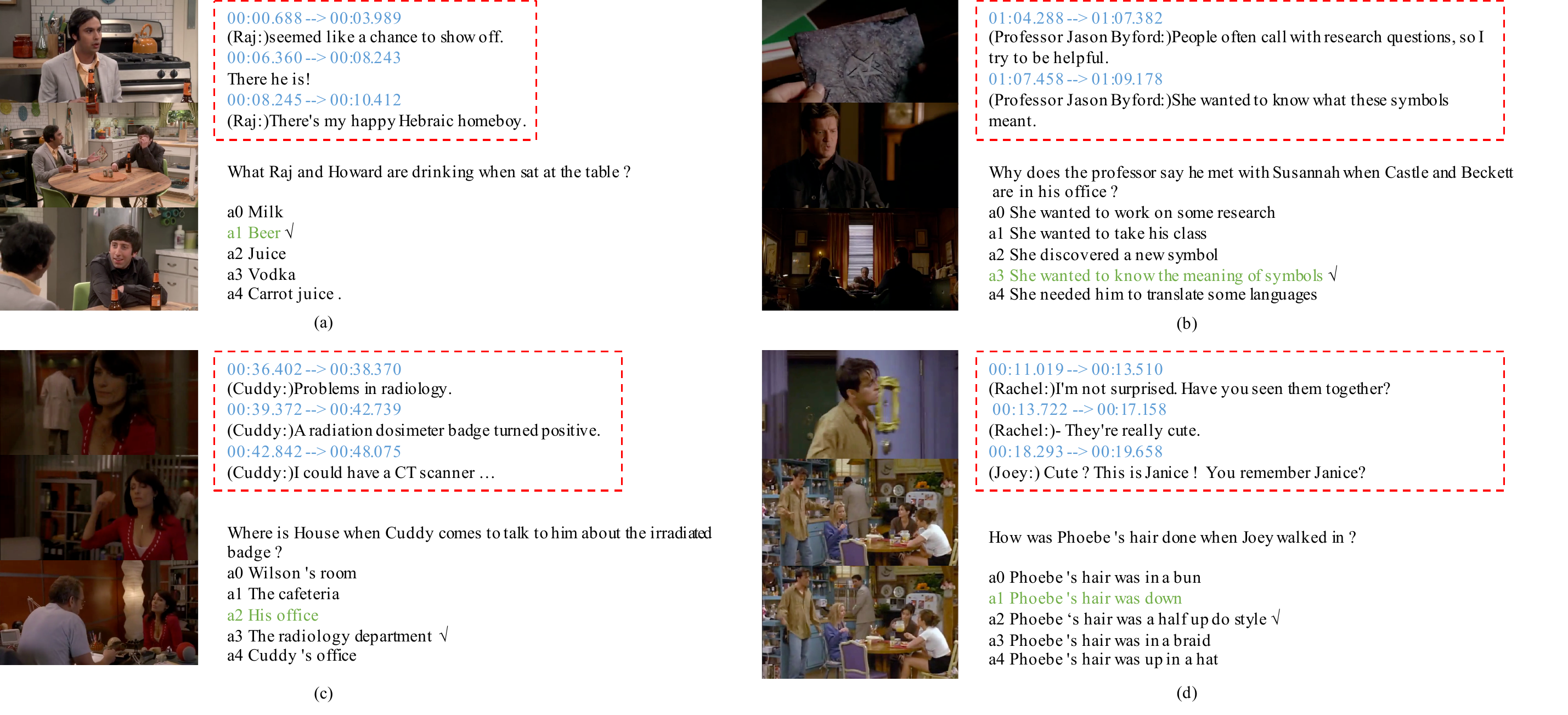}
  \vspace{-20pt}
  \caption{Example predictions from our best model. Top row shows correct predictions, bottom row shows failure cases. Ground truth answers are in green, and the model predictions are indicated by \checkmark. Best viewed in color.}
  \label{fig:pred_example}
  \vspace{-5pt}
\end{figure*}

\noindent\textbf{Comparison with Human Performance:} 
Human performance without timestamp annotation is shown in Table~\ref{tab:human_acc}. When using only questions (Table~\ref{tab:main_results} row 11), our model outperforms humans (43.50\% \textit{vs} 32.61\%) as it has access to all statistics of the questions and answers. 
When using videos or subtitles or both, humans perform significantly better than the models.

\noindent\textbf{Models with Timestamp Annotation:} 
Columns under \textit{w/o ts} and \textit{w/ ts} show a comparison between the same model using full-length videos/subtitles and using timestamp localized videos/subtitles. 
With timestamp annotation, the models perform consistently better than their counterpart without this information, indicating that localization is 
helpful for question answering.

\noindent\textbf{Accuracy for Different Question Types:} 
To gain further insight, we examined the accuracy of our models on different question types on the validation set (results in Table~\ref{tab:qtype_acc}), all models using timestamp annotation. Compared to S+Q model, S+V+Q models get the most improvements on ``what'' and ``where'' questions, indicating these questions require additional visual information. On the other hand, adding video features did not improve S+Q performance on questions relying more on textual reasoning, e.g., ``how'' questions.

\begin{table}[ht]
\centering
\scalebox{0.9}{
\begin{tabular}{l|l|c|c|cc}
\hline
&             & Video   & \multicolumn{2}{c}{Val Accuracy}  \\ \cline{4-5}
Method    &N.A. Src.    & Feature & w/o ts & w/ ts \\ \hline
V+Q       & Rand        &  cpt      & 84.64             & 85.01          \\ 
S+Q       & Rand        &  -        & 90.94             & 90.72             \\ 
S+V+Q     & Rand        &  cpt      & 91.55             & 92.00          \\ \hline
V+Q       & Human       &  cpt      & 43.03             & 45.03          \\ 
S+Q       & Human       &  -        & 62.99             & 65.15             \\ 
S+V+Q     & Human       &  cpt      & 64.70             & 67.70          \\ \hline
\end{tabular}
}
\caption{Accuracy on TVQA validation set with negative answers collected using different strategies. Negative Answer Source (N.A. Src.) indicates the collection method of the negative answers. Q = Question, S = Subtitle, V = Video, cpt = visual concept features, ts = timestamp annotation. All the experiments are conducted using the proposed multi-stream neural model. }
\label{tab:rand_negative}
\vspace{-5pt}
\end{table}

\noindent\textbf{Human-Written Negatives vs. Randomly-Sampled Negatives} For comparison, we create a new answer set by replacing the original human written negative answers with randomly sampled negative answers. To produce relevant negative answers, for each question, negatives are sampled (from the other QA pairs) within the same show. 
Results are shown in Table~\ref{tab:rand_negative}. 
Performance on randomly sampled negatives is much higher than that of human written negatives, indicating that human written negatives are more challenging.

\noindent\textbf{Qualitative Analysis:} 
Fig.~\ref{fig:pred_example} shows example predictions from our S+V+Q model (row 18) using full-length video and subtitle. 
Fig.~\ref{fig:pred_example}a and Fig.~\ref{fig:pred_example}b demonstrate its ability to solve both grounded visual questions and textual reasoning question. 
Bottom row shows two incorrect predictions. 
We found that wrong inferences are mainly due to incorrect language inferences and the model's lack of common sense knowledge. 
For example, Fig.~\ref{fig:pred_example}c, the characters are talking about radiology, the model is distracted to believe they are in the radiology department, while Fig.~\ref{fig:pred_example}d shows a case of questions that need common sense to answer, rather than simply textual or visual cues.

\section{Conclusion}

We presented the TVQA dataset, a large-scale, localized, compositional video question answering dataset.  
We also proposed two QA tasks (with/without timestamps) and provided baseline experiments as a benchmark for future comparison.
Our experiments show both visual and textual understanding are necessary for TVQA. 

There is still a significant gap between the proposed baselines and human performance on the QA accuracy. 
We hope this novel multimodal dataset and the baselines will encourage the community to develop stronger models in future work. 
To narrow the gap, one possible direction is to enhance the interactions between videos and subtitles to improve multimodal reasoning ability. 
Another direction is to exploit human-object relations in the video and subtitle, as we observe that a large number of questions involve such relations. 
Additionally, temporal reasoning is crucial for answering the TVQA questions. 
Thus, future work also includes integrating better temporal cues.

\section*{Acknowledgments}

We thank the anonymous reviewers for their helpful comments and discussions.
This research is supported by NSF Awards \#1633295, 1562098, 1405822 and a Google Faculty Research Award, Bloomberg Data Science Research Grant, and ARO-YIP Award \#W911NF-18-1-0336.
The views contained in this article are those of the authors and not of the funding agency.
\appendix
\section{Appendix}
\label{sec:supplemental}

\smallskip
\noindent{\bf Questions and Answers:}

\begin{figure}[h]
\centering
\scalebox{0.7}{  \includegraphics[width=\linewidth]{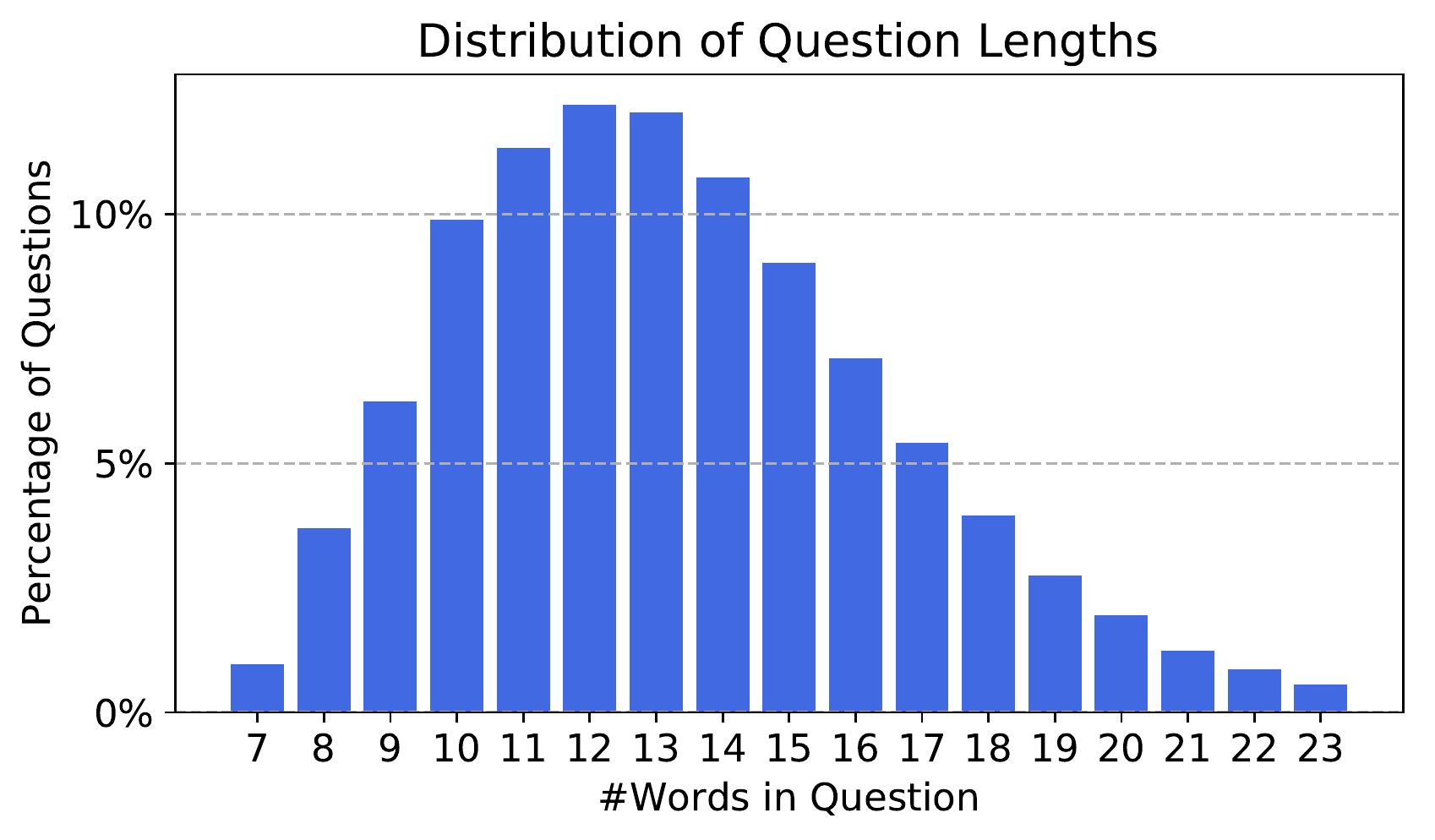}}
\vspace{-10pt}
  \caption{Distribution of question lengths.}
  \label{fig:q_len_dist}
 \vspace{-1pt}
\end{figure}

\begin{figure}[h]
\centering
\scalebox{0.5}{  \includegraphics[width=\linewidth]{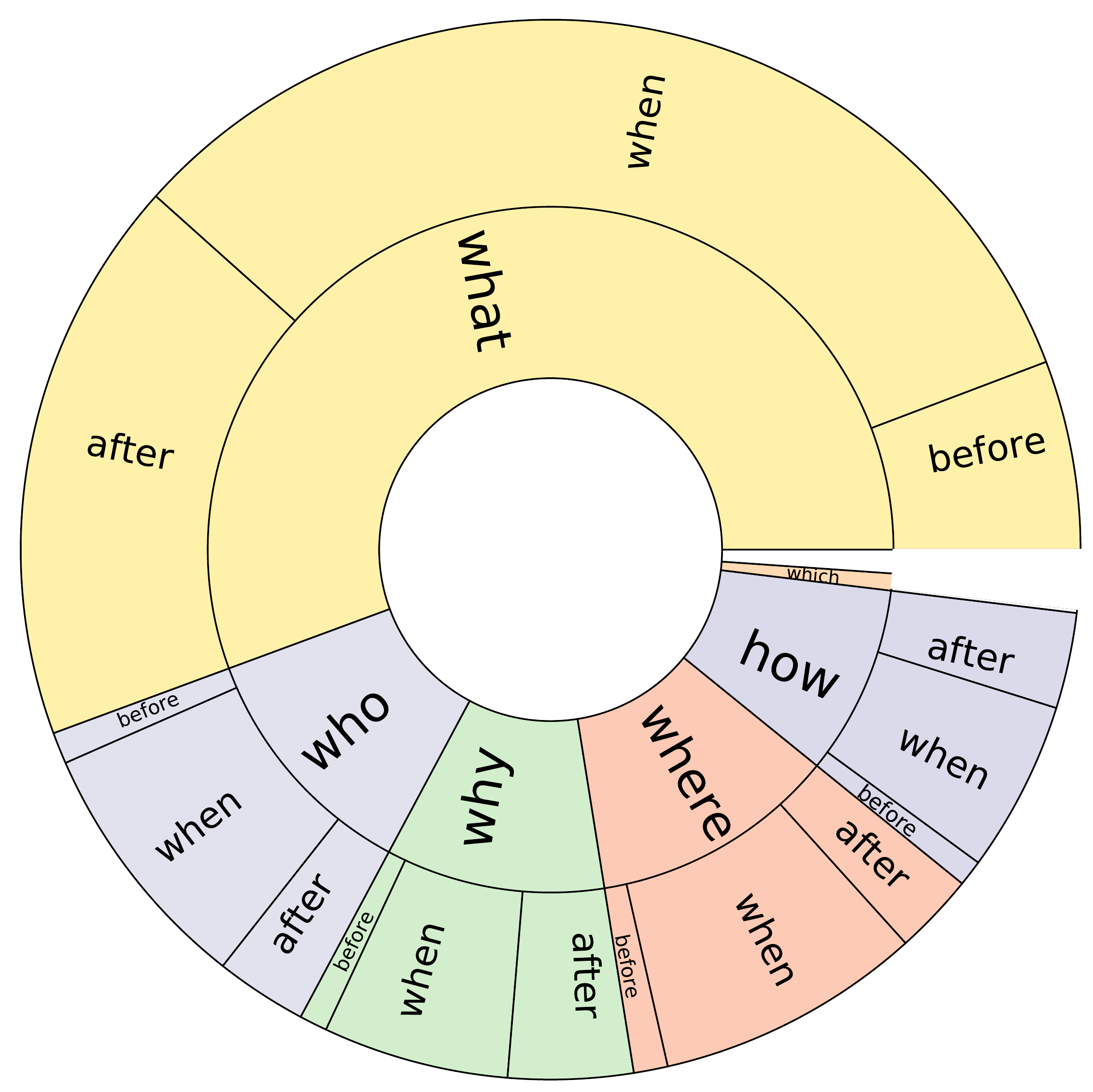}}
\vspace{-7pt}
  \caption{Distribution of questions by their first words and link words.}
  \label{fig:qtype_circle}
  \vspace{-1pt}
\end{figure}

\begin{figure}[ht!]
\centering
\scalebox{0.7}{  \includegraphics[width=\linewidth]{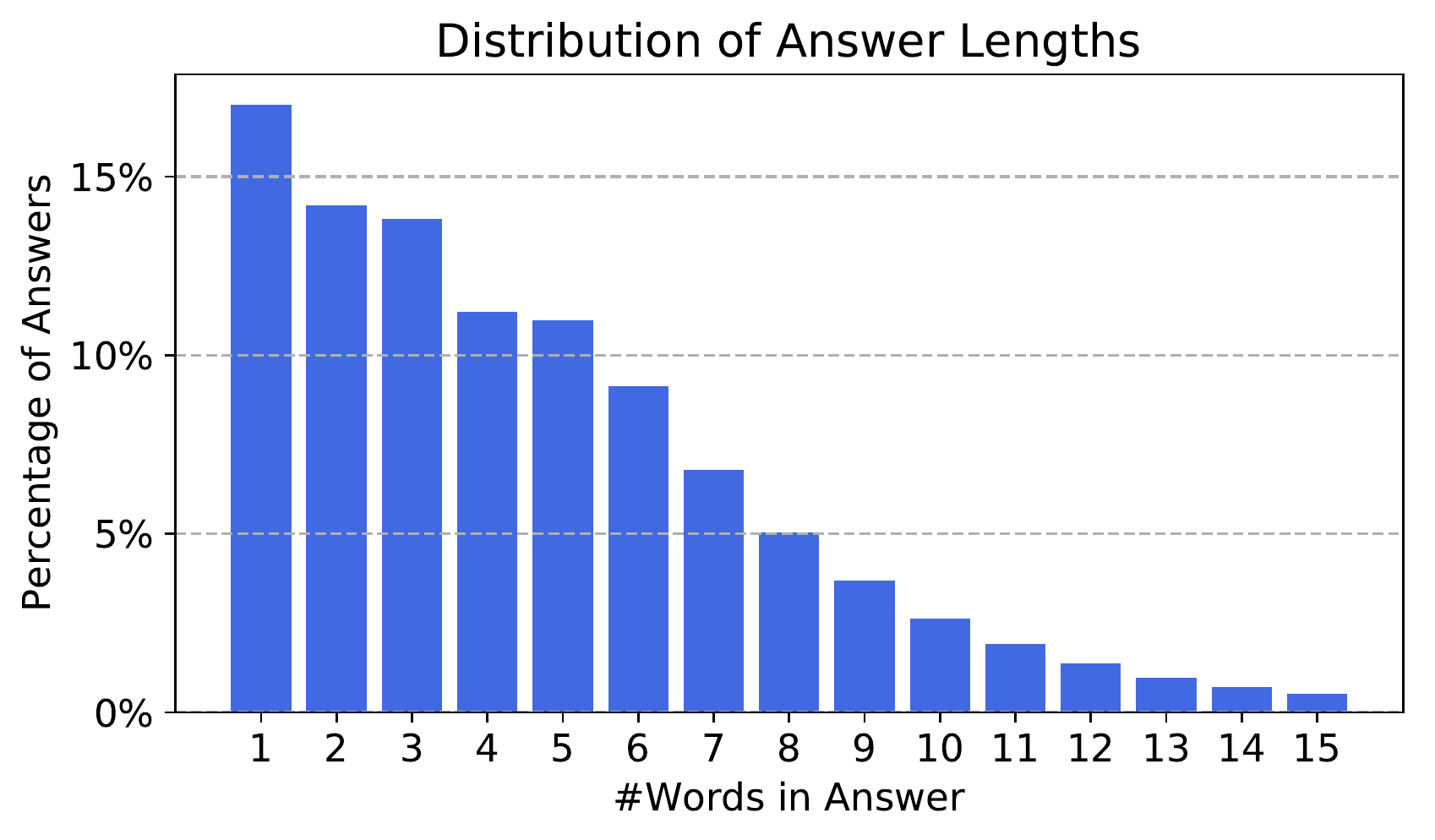}}
\vspace{-10pt}
  \caption{Distribution of answer lengths.}
  \label{fig:ans_len_dist}
\end{figure}

Fig.~\ref{fig:q_len_dist} shows the distribution of question lengths. Most of our questions are longer than 10 words. Fig.~\ref{fig:qtype_circle} shows a pie distribution of our first-part question words, as well as the associated second-part question words.
We see that ``what'' accounts for a large portion for the first part, i.e., ``what'' (Abstract, Object, Action), and ``when'' is more often used for moment localization in the second part.
Fig.~\ref{fig:ans_len_dist} shows the distribution of answer lengths. 
Note there is no peak answer length in TVQA, thus understanding the variable-length answers is also crucial for TVQA question-answering.

\begin{figure*}[h]
  \includegraphics[width=\linewidth]{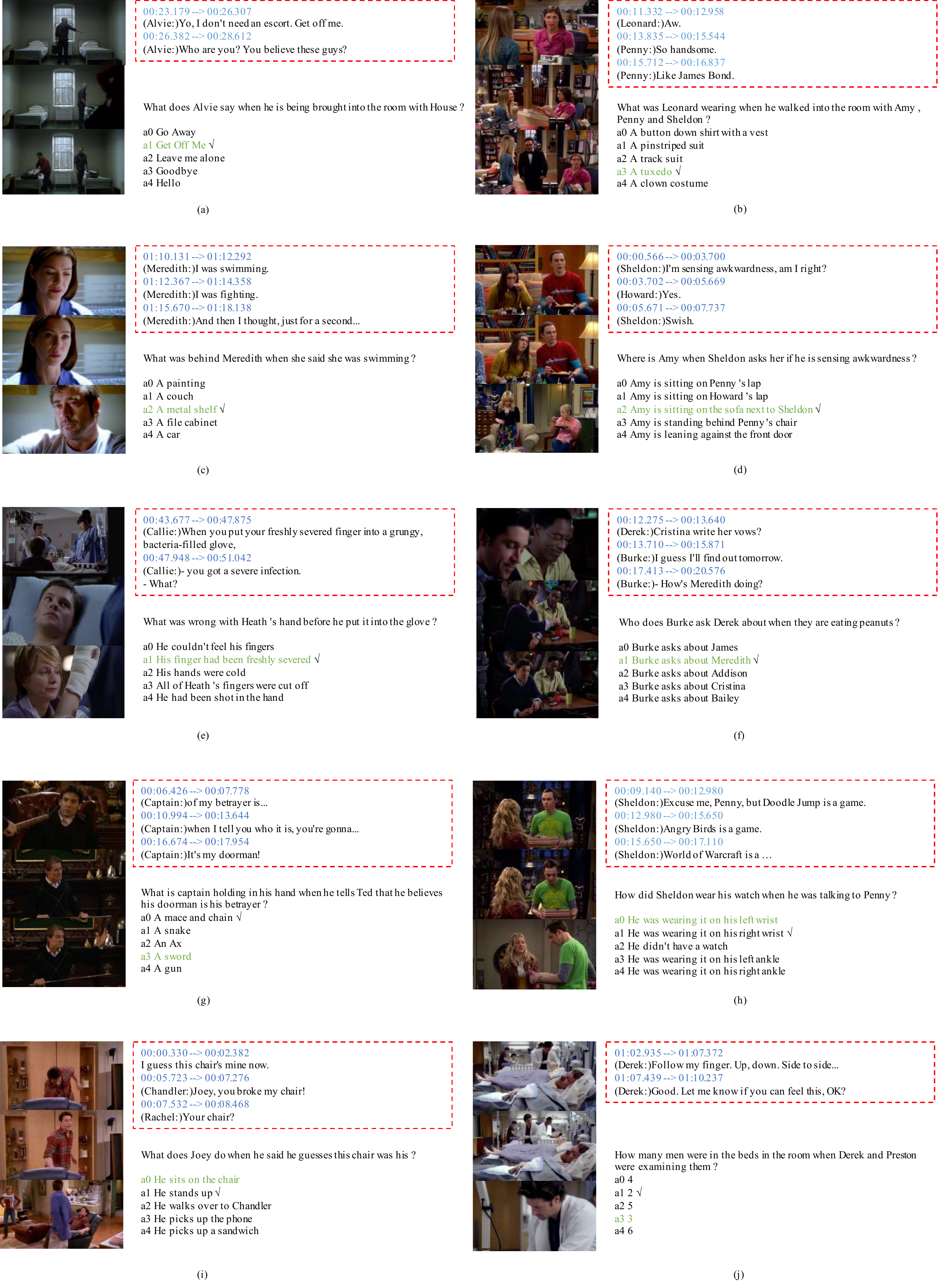}
  \caption{More examples from TVQA dataset, along with the predictions from our best model. Ground truth answers are highlighted in green, and model predictions are indicated by \checkmark. Best viewed in color.}
  \label{fig:examples}
\end{figure*}

\smallskip
\noindent{\bf More QA examples:}
We show more examples using our S+V+Q joint model trained on full-length video and subtitle in Fig.~\ref{fig:examples}, where the top three rows show the correct answers made by our model and the bottom two rows show the wrong ones.

\begin{table}[ht!]
\centering
\scalebox{0.65}{
\begin{tabular}{l|l}
\hline 
Character                & \multicolumn{1}{c}{Top unique nouns}                                                         \\ \hline
\multirow{2}{*}{Derek}     & Anna, MRI, today, tumor, black, desk,     \\
                         & car, area, trailer, procedure, ice       \\ \hline
\multirow{2}{*}{Cristina} & Colin, dress, leg, Doug, Deborah, wedding,             \\
                         & choker, restroom, lab, pad            \\ \hline
\multirow{2}{*}{Izzie}   & Cheyenne, check, Jilly, sweater, Heather,   \\
                         & bathroom, lunch, sheet, beast, condition   \\  \hline
\multirow{2}{*}{Burke}    & surgeon, valve, pole, cap, fishing, point,       \\
                         & situation, look, flower, pressure, breakfast    \\  \hline
\multirow{2}{*}{Meredith}   & Susan, thatcher, mom, counter, Dylan,   \\
                         & boyfriend, band, sister, step, Molly \\ \hline
\multirow{2}{*}{George}  & mackie, Sophie, conference, west, mercy, \\ 
                         & box, stairwell, pocket, bottle, morning\\ \hline
\multirow{2}{*}{Bailey}  & Viper, son, clinic, clipboard, gesture, binder, \\                         & Kelley, razor, scalpel, sex, piece, twin, Judy\\ \hline
\multirow{2}{*}{Addison}  & Naomi, ring, Sam, news, rose, Lisa, station,  \\                 & kid, ultrasound, pink, monitor, concern\\ \hline
\end{tabular}
}
\vspace{-5pt}
\caption{Top unique nouns for characters in Grey.}
\label{tab:grey_cooccur}
\end{table}

\begin{table}[ht!]
\centering
\scalebox{0.65}{
\begin{tabular}{l|l}
\hline 
Character                & \multicolumn{1}{c}{Top unique nouns}                                                         \\ \hline

\multirow{2}{*}{Ted}     & class, girlfriend, tattoo, movie, architect,    \\
                         &  Jen, word, Hammond, airport, student      \\ \hline
\multirow{2}{*}{Lily} & gift, purse, dress, kitchen, water, card, tree,            \\
                         & plan, Judy, credit, mouth, paint, bedroom             \\ \hline
\multirow{2}{*}{Barney}   & James, Jerry, Nora, suit, brother, Bob,   \\                    & Sam, strip, cigar, way, fist, street, tie       \\  \hline
\multirow{2}{*}{Marshall}    & sandwich, brad, burger, Jeff, money,       \\
                         & song, work, Marvin, laptop, school,      \\  \hline
\multirow{2}{*}{Zoey}   & landmark, thanksgiving, cousin, tape,   \\
                         & husband, spot, bit, turkey, text, recording  \\ 
                         \hline
\end{tabular}
}
\vspace{-5pt}
\caption{Top unique nouns for characters in HIMYM.}
\label{tab:himym_cooccur}
\end{table}

\smallskip
\noindent{\bf Top unique nouns for characters:}
We show the top unique unique nouns for characters for Grey and HIMYM in Table~\ref{tab:grey_cooccur}~\ref{tab:himym_cooccur}.

\clearpage
\clearpage

\bibliography{emnlp2018}
\bibliographystyle{acl_natbib_nourl}

\end{document}